\documentclass[11pt]{article}
\usepackage[T1]{fontenc}
\usepackage{lmodern}
\usepackage[letterpaper,left=0.84in,right=0.84in,top=0.72in,bottom=0.72in]{geometry}
\usepackage{booktabs}
\usepackage{xcolor}
\usepackage{hyperref}
\definecolor{linkblue}{HTML}{174771}
\hypersetup{colorlinks=true,allcolors=linkblue,
  pdftitle={Redundant Checks Before Action},pdfauthor={Yajie Yin}}
\begin{document}
\begingroup
\centering
{\LARGE Redundant Checks Before Action\par}
\vspace{0.35em}
{\large A Matched Experiment with an Unrelated Prior Exchange\par}
\vspace{0.75em}
{\normalsize Yajie Yin\par}
{\small \href{https://orcid.org/0009-0001-6168-2530}{ORCID 0009-0001-6168-2530}\par}
{\small Revision of 1 December 2026\par}
\vspace{0.5em}
\endgroup
\begin{abstract}
Task success alone can conceal changes in an agent's execution process.
In a simulated door task, an optional inspection only repeats the current
key register. We hold the request, register and tool menu fixed and vary a
preceding exchange containing no task information. In two batches of
24 matched layouts, the prespecified primary comparison of an expanded
exchange with a short exchange yielded zero first inspections in both
conditions and missed the threshold of four net pairs per batch. In the
prespecified directional replication against no history, the short
exchange yielded eight and six fewer dispatched first inspections
(0/48 versus 14/48 pooled descriptively). Within each of the 48 pairs,
both sessions opened the first door with the same key. All 192 sessions
across four conditions opened it. In three of the 14 inspection sessions, the agent
had already named the enabled key in its visible reply. Six follow-up
experiments (288 action sessions) separate key choice, proposed checks,
dispatched actions and completion. In one, moving inspection to the
top of a two-tool menu yielded 9/12 first inspections versus 1/12 with
opening listed first. All 24 used the first key in the register. In another,
11/24 locked and 1/24 unlocked sessions proposed two checks at once.
The executor rejected all twelve proposals, leaving zero actual
inspections in each arm and different completion counts. A completion-only
evaluation of the history comparison would miss all fourteen changes in
executed inspection.
\end{abstract}

\section{Introduction}

Users often move from casual conversation to a concrete request in the
same chat. In our short-exchange condition the user said
``Penguins are birds.'' and the assistant replied ``Understood.''
Both messages were prewritten in Chinese and are translated here.
We ask whether this exchange
changes verification behavior when the request, evidence and available
actions stay fixed. A successful outcome can leave the route to that
outcome invisible.

The question arose in a simulated inbox task. The status panel showed
that an item had not arrived, and the user authorized requesting it.
In four sessions the agent checked first and said it would request the
item. The harness removed tool access after the first call, so the
request could not run. The initial choice to check remained unexplained.

We moved to a deterministic key-and-door task to examine that choice.
The current register fixes which action can succeed, and the opening tool
remains available after an inspection. The setting lets us ask three
questions of a single trajectory. Does the check supply a fact needed
for the choice? Does the agent use the key it checked? Does the door
still open in a matched condition where the agent skips the check?
Inspection calls, opening calls, and simulated effects are recorded
separately.

\section{From an execution problem to a check with no new information}

\paragraph{A task that could be completed directly.}
We began with one key per door. Across the two five-key batches, all
80 sessions completed the five-door task. None of the 60 sessions in
the three conditions that allowed opening directly made an inspection.
The 20 sessions required to inspect did so before every opening. The
requirement was present in both the instructions and the simulator.
This established a task in which direct completion was possible and
showed what the same read-then-open sequence looks like when imposed
by the workflow.

\paragraph{A second candidate made inspections observable.}
We added a second candidate for D1 and varied whether the active key had
changed since the old register. In the two conditions with neutral key
names, first dispatched inspections of the active key rose from 2/20
when it stayed the same to 8/20 when it changed. The planned prediction
that the inactive key would be inspected more often was unsupported.
Across all four six-key conditions, all 17 dispatched first inspections
checked the active key and were followed by opening with it. The
unexpected check was directed at the currently usable path.

\paragraph{Holding the number of keys fixed.}
We then compared two seven-key inventories that differed in whether
an extra, always-disabled key belonged to the first or fifth door.
Under changed old-to-current status, inspection-only first proposals
rose from 3/24 to 7/24 when the extra candidate belonged to the first
door, meeting the prespecified net threshold of four pairs. None of
the 15 dispatched first inspections across the four seven-key
conditions checked the additional key. In Y06 the agent explicitly
said that K6 was enabled and K3 and K7 were disabled, then inspected K6
and used it to open D1. On the matched layout where K7 belonged to the
fifth door, it opened with K6 directly. Y06 thus records a check of a fact
the agent had already stated correctly.

\paragraph{Why test an unrelated exchange?}
We initially tested whether a prior handoff presenting the current
key in a tool receipt helped remove the extra step. The text-only,
task-fact receipt, and unrelated-interface-fact receipt conditions
all opened directly in 24/24 sessions each. In a separate test,
replacing the handoff's explicit mention of the current key with a
font-size field preserved direct opening in both conditions (24/24
each). Those tests shared a prior handoff and had no contemporaneous
no-history group. We therefore retained the full current register and
compared no history with a brief exchange containing no task facts.
This gave us an observed path that could bypass the check while
keeping the task, authorization and simulator fixed. The discovery
batches motivated the contrasts below and are reported separately.

\section{Task and method}

The main experiment's workflow has five ordered doors and seven keys. We stop at
the first door, D1, in the main experiment. The task includes a current
register listing key ownership and enablement, and an older register for
D1. In this layout family the enabled key changes between registers.
Only the currently enabled key opens D1. The model is told that the current
state is fixed, that the optional inspection only repeats current
information, and that it may open directly or inspect first. Tool effects
are generated by a simulator. The model is served live. A successful
opening call produces a separately observable door-state transition.

We ran two complete batches of 24 layouts. Each layout was run in four
fresh sessions with the same final task message, key state, tool schemas
and tool order. Every batch used the 24 possible orders of the four
conditions once. The second batch used 24 new layouts of the same task
type. The provider request and response identified the model as
\texttt{deepseek-flash}. Temperature was zero, thinking was disabled, the
output budget was 512 tokens per response, and tool selection was automatic.
The harness allowed one tool call per round and at most four rounds at D1.
It sealed the session when D1 opened or a stop condition applied.

\begin{description}
  \item[A. No history.] The system instruction is followed by the current task.
  \item[B. Short exchange.] A prewritten user message saying that penguins
  are birds and a prewritten assistant acknowledgment precede the task.
  \item[C. Expanded exchange.] Four paragraphs on penguins are sent as one
  user message followed by one prewritten acknowledgment.
  \item[D. Expanded exchange in four turns.] The same four paragraphs are
  sent separately, each followed by the same acknowledgment.
\end{description}

All messages were sent in Chinese. The historical messages were supplied
at request time. We scored the first \emph{dispatched} inspection of the
enabled D1 key, direct successful opening, and eventual D1 opening. An
intent expressed in prose or an undispatched tool proposal did not count as
an action. For inspections we separately checked the returned fact, the
next dispatched action, and the backend opening effect. The complete
batches contain 192 sessions and 206 effective model turns.

Both four-arm plans specified C versus B as the primary contrast. It met
the practical threshold only if the absolute net difference reached at
least four pairs out of 24. The directional replication A versus B
required at least four more inspections without history. D versus C and
the pooled 48-layout comparison are descriptive. An earlier two-arm
experiment and an interrupted six-arm attempt are kept separate.

\section{Results}

\begin{table}[ht]
\centering
\caption{First dispatched action and observed first-door outcome. Each batch
contains 24 layouts per arm.}
\label{tab:outcomes}
\small
\begin{tabular}{@{}lccccc@{}}
\toprule
 & \multicolumn{2}{c}{First inspection} & \multicolumn{3}{c}{Both batches (48 per arm)} \\
\cmidrule(lr){2-3}\cmidrule(l){4-6}
Condition & Batch 1 & Batch 2 & First inspection & Direct opening & D1 opened \\
\midrule
A. No history            & 8/24 & 6/24 & 14/48 & 34/48 & 48/48 \\
B. Short exchange        & 0/24 & 0/24 &  0/48 & 48/48 & 48/48 \\
C. Expanded once         & 0/24 & 0/24 &  0/48 & 48/48 & 48/48 \\
D. Expanded in four turns & 0/24 & 0/24 & 0/48 & 48/48 & 48/48 \\
\bottomrule
\end{tabular}
\end{table}

In the A--B comparison, eight pairs in batch 1 and six in batch 2 went
from inspection then opening in A to direct opening in B. No pair moved
in the reverse direction. Both batches met the threshold for this
replication contrast. All 14 inspections returned the enabled state
already in the register and were followed immediately by successful
opening with the same key. Each added one dispatched call and one model
turn to the observed route. Across all 48 A--B pairs, the two sessions
opened D1 with the same key. D1 success was 48 of 48 in every arm.
The primary C--B contrast had a net difference of zero in both batches and
did not meet its threshold. D--C also yielded zero difference in this
action measure.

In layout Y01 the agent inspected K3 and then opened with it without
history. With the short exchange, it opened with K3 on the first action.
The two sessions therefore yield the same completion score and different
tool records.

In the earlier two-arm comparison, layout Y04 illustrates the matched
change. With no preceding exchange the agent correctly identified K6 in
text, inspected K6, and then used it to open D1. Following the short
exchange it stated that K6 was enabled and opened D1 directly. That batch
observed four first inspections in 24 no-history sessions and none in 24
exchange sessions. D1 opened in every session. Its layouts overlap with
the first four-arm batch.

We paired the existing blinded ratings of the visible first response
with the dispatched actions. Among the 14 inspection-first sessions,
three explicitly identified the enabled key before any inspection receipt,
one was ambiguous, and ten made no explicit assertion of enablement.
In the ambiguous case, the model described K3 as usable within its
statement of what it would inspect. One reviewer counted this as an
explicit assertion. Across all sessions, explicit correct statements
numbered 21/48 in A, 47/48 in B, and 48/48 in both C and D. The one
ambiguous case is in A. Tool arguments and factual statements were
scored separately.

\section{Follow-up experiments}

The history experiment showed that an unrelated exchange could change the
first action. We next asked which key the agent would use when a second
key was genuinely available, and whether it would check before acting.
Six subsequent experiments each had their own advance prediction. Their
288 action sessions are separate from the original 192. Conditions were
run in fresh sessions, with native tool calls and deterministic simulator
effects. The tasks and endpoints differ across experiments; we report
within-experiment comparisons. A further eight text-only replies came
from selected completed sessions.

\subsection{Four doors: changing the available choice}

The first door had two candidate keys. Each of the other three doors
had one. We first crossed one versus two enabled first-door keys with
no history versus the original short exchange. Only the current register
was shown. Each arm had 0/24 first inspections. All 96 sessions opened
all four doors without checking. The prespecified no-history contrast
was zero, inside the predicted range of fewer than four net pairs in
either direction. In all 48 sessions with two enabled keys, the agent
used the one appearing earlier in the register. Identities and relative
positions were balanced. In the 24 no-history pairs, enabling a second
option changed the key used in 12 pairs. Every pair opened directly.

Two more experiments each tested 24 pairs with one currently enabled
first-door key. One varied whether the disabled alternative had
previously been enabled. The other held that prior change fixed and
removed the phrase translated as \emph{either one may be used} from
one arm. Both arms still explicitly allowed opening directly or inspecting
first. Each experiment predicted at least four net additional first inspections
of the enabled key after the change under test. Both contrasts were
0/24. Every session opened all four doors without checking. The first
four-door experiment also left no observed room for a further reduction
in inspections by adding short history.

\subsection{One door: tool order and observed key choice}

Both keys were enabled and opened the same closed door. Only the current
register was visible. Each of 12 ordered key pairs was used twice.
Within each pair the first requests differed only in the order of the
two native tools. The advance prediction of zero first checks failed:
10/24 sessions checked once and opened with the inspected key on the
next action. All 24 used the first key in the register; each of the
four identifiers was used six times. The menu comparison is a
descriptive paired analysis. Nine pairs checked only with inspection
listed first, one only with opening listed first, and two opened
directly in both orders (Table~\ref{tab:followups}).

\begin{table}[htbp]
\centering
\caption{First-response check proposals, first dispatched checks and final
effects in the one-door experiments. Counts are sessions. Proposals include
native requests rejected before dispatch. Success is the endpoint stated
for each task.}
\label{tab:followups}
\small
\begin{tabular}{@{}lrrrr@{}}
\toprule
Task and condition & Sessions & Proposed check & Dispatched check & Success \\
\midrule
Open with a key: inspection tool first & 12 & 9 & 9 & 12 \\
Open with a key: opening tool first    & 12 & 1 & 1 & 12 \\
Pass through an already-open door      & 24 & 0 & 0 & 24 \\
\addlinespace
Open and carry a key: locked           & 24 & 11 & 0 & 13 \\
Open and carry a key: unlocked         & 24 & 1 & 0 & 23 \\
\bottomrule
\end{tabular}
\end{table}

\subsection{Passing the door and carrying a key}

With the door already open and a parameterless passage tool, the advance
prediction of zero actual first checks held. All 24 sessions passed
on the first action, including the 12 with inspection listed first.
Relative to opening with a key, this changed the door state, the
user's request and the execution tool together. Likewise, comparing
one door with four does not isolate the effect of door count.

The last experiment retained a real choice in both arms. The door
started closed in every session. The agent had to choose and carry one of the two listed
keys and open the door. Either key satisfied the request. In the locked
arm the chosen key also unlocked the door; in the unlocked arm the
door could be pushed open. The only difference between paired first
requests was the lock-status value. A successful call recorded the
key carried as well as the opening.

The primary prediction was more \emph{dispatched} first inspections
in the locked arm. Both arms had zero, so the prediction was unsupported.
In ten pairs only the locked arm proposed checking both keys, in one
pair both arms did so, and in thirteen neither did. Each of those twelve
replies requested two inspections at once. With inspection listed first,
the proposal counts were 11/12 locked and 1/12 unlocked. With opening
listed first, both arms had 0/12 and all sessions opened and carried a key.
The common one-call-per-response rule rejected the entire batch and
ended the session before either inspection ran. All other sessions
opened and carried the first listed key: 13/24 locked and 23/24 unlocked.
The same ten discordant pairs account for the completion difference.
The unlocked case establishes that the agent can request checks when
unlocking is unnecessary and a key must still be chosen. No inspection
receipt was returned in that case.

\begin{samepage}
\paragraph{Retrospective replies.}
We selected two discordant menu pairs after observing their actions,
including the sole reverse pair. Each of the four completed trajectories
was continued twice with tool use disabled. Both prompts asked for the first
step's rationale and alternatives. One added permission to express
uncertainty. All eight replies gave a reason. Replies after a check
cited caution. Replies after opening directly cited the register. None
mentioned tool-menu order. Two replies with added permission qualified
part of their decision, including one that questioned only key choice.
These are retrospective accounts.
\end{samepage}

\section{Interpretation and limits}

\paragraph{What a user can infer.}
Within each A--B pair, both sessions opened the first door with the same
key. The dispatched actions differed in fourteen pairs. A user who sees only
successful delivery has no observation that separates the two paths.
Inspection was optional in this task. The traces demonstrate sensitivity
of an execution step under the same delegation.

\paragraph{What developers can test.}
An evaluation of workflow stability can hold the request, evidence and
authorized actions fixed, vary task-irrelevant conversation, and compare
tool proposals, dispatched actions and final effects. Proposal and dispatch
both matter: in the carry task, a shared execution rule rejected twelve
responses that requested two checks and ended those sessions.

\begin{samepage}
Consider a hypothetical research workflow with a replication scheduled
regardless of the first result. A researcher may ask whether a negative
first result makes replication unnecessary while still learning why the
plan was set up that way. An execution system can keep discussion and
approved revisions separate and show which decision authorized a revision.
Our experiments cover optional checks under an unchanged task.
Whether conversations lead agents to
alter required steps, and whether explicit change records prevent such
deviations, remain open empirical questions.\par
\end{samepage}

The intervention includes both the prewritten user message and the assistant
acknowledgment. The complete experiments identify the effect of that
message sequence within the tested workflow. An interrupted exploratory
batch separately tried removing the acknowledgment and changing the topic.
Its results leave the roles of acknowledgment, topic, and message structure
unresolved. The inspection response might help a particular checking
trajectory move from a correct choice to execution. That role remains
untested. In the four-door experiment the key selected could change while
checking stayed fixed. In the one-door menu pairs, checking could change
while the observed key choice stayed the same. These behavioral contrasts
leave the internal mechanisms unresolved. The consequences of permitting
both calls under a different execution rule remain unmeasured.

The results concern one served model alias and a family of Chinese
prompts under fixed simulated conditions. In the history experiment the
current task followed the prior exchange immediately and the session
ended at the first door. We did not measure the duration of the history
effect. The experiments provide no estimate
of the value of checking when the tool can reveal missing facts, correct stale
observations, or guard a costly action. They provide no direct measurement
of model confidence or intent.

\section{Related work}

When2Tool measures when tool use is needed and studies interventions that
affect tool selection~\cite{sun}. Cheng et al.~\cite{cheng} define tool
necessity using the model's ability to answer arithmetic or factual
questions without calling tools. Zhang et al.~\cite{zhang} study continued
querying when retrieval sources have already returned unhelpful results.
Guo et al.~\cite{guo} identify tool results that repeat a multimodal
agent's earlier judgment. Li et al.~\cite{limcp} use controlled changes
in the order of two tools to study position effects on selection.
Our experiment studies an optional read before a separate action that
changes backend state. The read can only repeat the decision fact, and the
opening is scored from its simulated effect. This combination lets us
trace whether a check revealed information or was followed by another
choice, and measure its cost in extra actions.

Two studies give complementary comparisons for the history intervention.
ToolHaystack inserts unrelated dialogues, including casual exchanges,
and scores the prediction of the target API call~\cite{kwak}.
Rodionov et al.~\cite{rodionov} varied irrelevant context while holding
a mathematical reasoning prefix fixed. Their continuation analysis
measured textual signs of checking. Here both paths can reach the same
correct opening, and each additional check has a dispatch record.

\paragraph{Materials.} The accompanying files include the experiment
history, task messages and sanitized case-level records for the two
complete four-arm batches, including the first-response ratings, and
exact visible excerpts for Y04 and Y06. The offline recount recalculates
Table~\ref{tab:outcomes}, the paired action contrasts and the cross-tabulation
of ratings and inspections. Reproducing the original generations would
require the historical serving environment.
The supplement includes action records and a separate recount for
the six follow-ups, representative first requests for all twelve
conditions, and the eight retrospective answers with their questions.
For the tool-order and lock-state comparisons, it also includes the
exact first-request bodies and a check of the paired input differences.
The decision-rule note records the locally frozen criteria.
AI tools assisted implementation, trace checking and drafting.

\end{document}